\documentclass{article}
\usepackage{spconf,amsmath,graphicx}
\usepackage{subfigure}
\usepackage{makecell,multirow,diagbox}
\usepackage{xcolor}
\usepackage{comment}
\usepackage[pagebackref=true,breaklinks=true,letterpaper=true,colorlinks,bookmarks=false]{hyperref}
\ninept
\usepackage{caption}
\AtBeginDocument{
	\setlength{\abovedisplayskip}{0pt}
	\setlength{\belowdisplayskip}{0pt}
}

\usepackage{lipsum}
\newcommand\blfootnote[1]{%
	\begingroup
	\renewcommand\thefootnote{}\footnote{#1}%
	\addtocounter{footnote}{-1}%
	\endgroup
}

\title{DarkShot: Lighting Dark Images with Low-Compute and High-Quality}
\name{Jiazhang Zheng \qquad Lei Li$^{\dagger}$ \qquad Qiuping Liao$^{\dagger}$ \qquad Cheng Li \qquad Li Li \qquad Yangxing Liu$^{\ast}$}
\address{TCL Corporate Research, Wuhan 430000, P.R. China}
\usepackage{fancyhdr}
\begin{document}
%
\maketitle
\begin{abstract}
Nighttime photography encounters escalating challenges in extremely low-light conditions, primarily attributable to the ultra-low signal-to-noise ratio.
For real-world deployment, a practical solution must not only produce visually appealing results but also require minimal computation.
However, most existing methods are either focused on improving restoration performance or employ lightweight models at the cost of quality.
This paper proposes a lightweight network that outperforms existing state-of-the-art (SOTA) methods in low-light enhancement tasks while minimizing computation.
The proposed network incorporates Siamese Self-Attention Block (SSAB) and Skip-Channel Attention (SCA) modules, which enhance the model's capacity to aggregate global information and are well-suited for high-resolution images.
Additionally, based on our analysis of the low-light image restoration process, we propose a Two-Stage Framework that achieves superior results.
Our model can restore a UHD 4K resolution image with minimal computation while keeping SOTA restoration quality.
\end{abstract}

\section{Introduction}
\label{sec:intro}
Low-light imaging has gained significant attention in recent years, owing to its wide range of applications across diverse domains such as night-time surveillance, target detection, and night photography. 
Despite technological progress, cameras still struggle in very low-light situations, leading to issues like color distortion and noise.
Several photography techniques can be utilized to improve image quality in low-light condition. but each of them has inherent limitations.
For example, long exposure time creates motion blur, high ISO settings amplifies noise, and using flash can cause uneven lighting.

In recent years, deep learning-based methods have been proven their effectiveness in the domain of low-light imaging. 
Chen et al.\cite{sid_chen2018_sid} collected the SID dataset and succeeded in restoring extreme low-light raw images captured in near zero lux conditions (0.1-5 lux).
To achieve this, they developed an end-to-end training pipeline and utilized the U-net architecture to learn how to enhance dark images. This method eliminated the error accumulation associated with traditional camera processing pipelines and produced superior results.
Since then, numerous deep learning architectures have been proposed to enhance the quality of dark images \cite{did_maharjan2019did, lcd_xu2020lcd, abandoning_dong2022abandoning, sgn2019gu, llpack_lamba2020LLPackNet, red_lambaCvpr2021Red}.

However, these advancements to enhance image quality are not necessarily making networks more efficient as they always come at a cost: high computational resources. 
For example, the current state-of-the-art (SOTA) models DBLE\cite{abandoning_dong2022abandoning} and LDC\cite{lcd_xu2020lcd} require 1162.51 GMACs and over 2000 GMACs (Giga Multiply-Accumulate Operations) respectively, to restore a single raw image from the SID dataset \cite{sid_chen2018_sid}. 
To reduce computation complexity, LLPackNet\cite{llpack_lamba2020LLPackNet} and RED\cite{red_lambaCvpr2021Red} designed lightweight models at a level of 83.46 GMACs and 59.8 GMACs but resulted in slightly worse reconstruction quality. 

\begin{figure*}
	\begin{center}
		\includegraphics[width=0.9\linewidth]{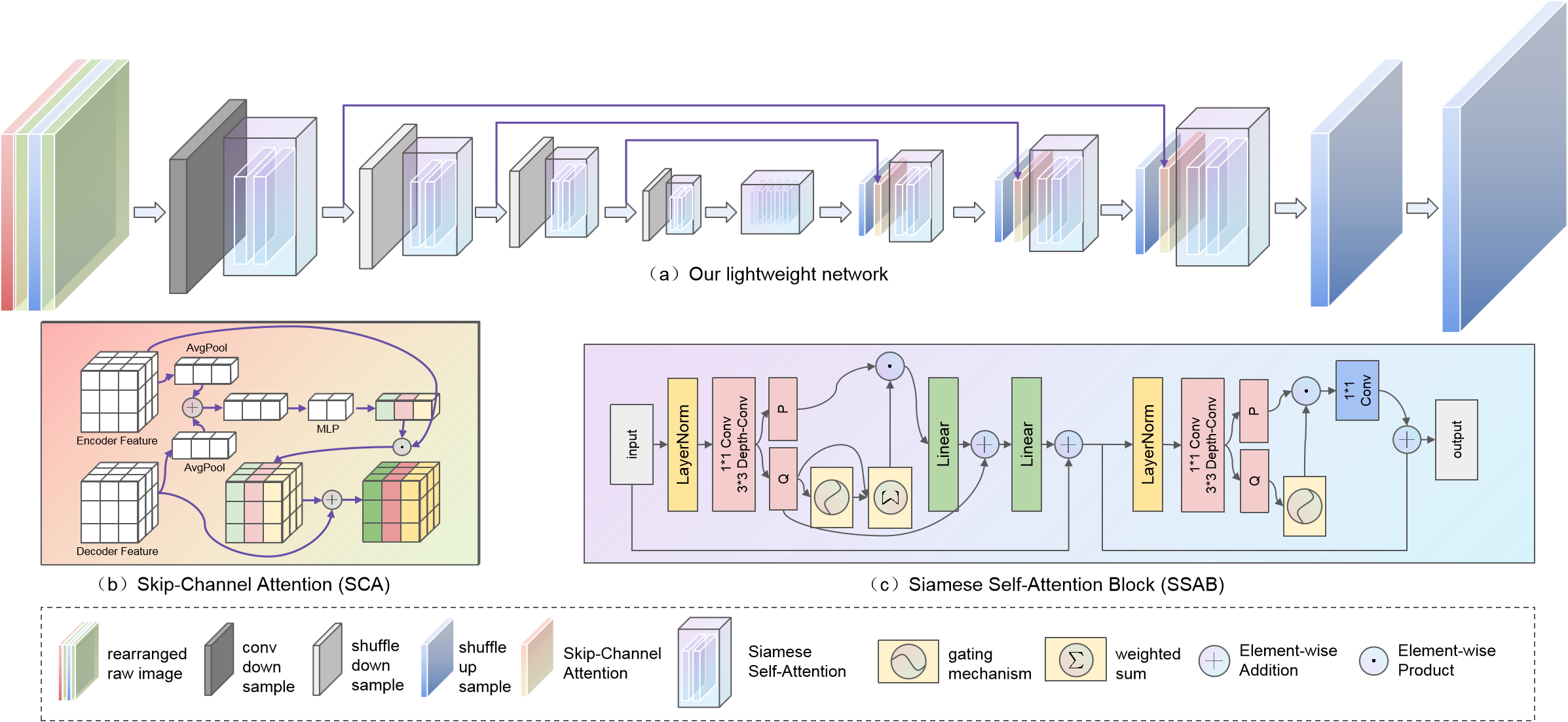}
	\end{center}
	\caption{The proposed network architecture}
	\label{fig:fun-net}
	
\end{figure*}

In this paper, we propose a lightweight network architecture that achieves the best restoration performance while requiring significantly less computational resources.
Illustrated in Fig.\ref{fig:fun-net}, we adopt a UNet-like design, incorporating our novel Siamese Self-Attention Block (SSAB) and Skip-Channel Attention (SCA). Specifically, SCA enhances the network's ability to effectively integrate salient features across different layers, making it ideal for wider utilization.

Self-attention is a critical component in Vision Transformers (VIT), which have achieved state-of-the-art performance in numerous visual tasks.\cite{dosovitskiy2020vit, carion2020detr, liang2021swinir, liu2021swin}.
VIT can effectively gather global information from feature maps, but its computational and memory cost grow quadratically with input size, rendering it impractical for high-resolution images. 
To address these issues, we propose the Siamese Self-Attention Block, which exhibits linear computational complexity with respect to input size and can be applied to feature maps of various resolutions.
Initially, the input features are split into two distinct segments. 
One of these segments is employed to efficiently compute attention weights through a saliency-gate mechanism.  Subsequently, the computed global context is merged with other segments to facilitate local-global interactions.
Notably, our approach does not require the calculation of pairwise interactions between input visual elements in the common dot-product attention, which is the cause of quadratic computational complexity.

Amplifying raw images with the desired ratio is a commonly used solution for brightening dark images\cite{did_maharjan2019did, lcd_xu2020lcd, abandoning_dong2022abandoning, sgn2019gu, llpack_lamba2020LLPackNet, red_lambaCvpr2021Red}.
However, this approach introduces a significant amount of noise that ultimately degrades the reconstruction quality. 
It is challenging to attain a trade-off between brightness enhancement and noise reduction in a single model due to their domain conflicts. 
These observation provides insights for us to decouple mappings of different tasks, resulting in the design of a two-stage model that produces improved results.

Our contributions can be summarized as follows:
\blfootnote{$^{\ast}$Corresponding author (yangxing.liu@tcl.com). $^{\dagger}$Equal contribution.}

{\bf Lightweight.} A lightweight network is proposed to enhance the quality of extremely dark, high resolution images using minimal computational resources. 
Our attention mechanism and lightweight model lower the computational complexity enough for the restoration of 4K images while achieving state-of-the-art results.

{\bf SSAB.} We propose the novel and efficient saliency-gate mechanism to compute interactions between global context and input visual elements. The dimension of global context is independent of and much smaller than input spatial size, ensuring linear computational complexity w.r.t input's spatial size.

{\bf SCA.} Previous methods that solely compute attention on encoder features suffer from ambiguity in fusing cross-layer information. SCA addresses this issue by incorporating guidance from decoder, enabling more effective fusion from different layers.

{\bf Two-stage framework.} We introduce the decoupling of noise and brightness tasks in low-light raw image enhancement. 
Leveraging the unique attributes in both the RAW and sRGB domains, our approach achieves strong performance while maintaining simplicity. 

\clearpage
\pagestyle{plain} 

\section{PROPOSED METHOD}
\label{sec:Approach}

\subsection{Network Architecture}\label{network}

{\bf Base Network.} 
A U-shape structure is employed to build the base network, which has proven its effectiveness in low-light image enhancement \cite{sid_chen2018_sid, abandoning_dong2022abandoning, llpack_lamba2020LLPackNet, red_lambaCvpr2021Red}. 
Such structure enables the efficient merging of hierarchical multi-scale information by combining low-level features with high-level features.
Furthermore, Our network adopts more efficient pixel-shuffle and pixel-unshuffle sampling methods from image super-resolution task\cite{shi2016espcn}.

{\bf Skip-Channel Attention.}
Skip connections are commonly utilized in U-shaped structures to merge information from shallow and deep layers\cite{hdrnet_eilertsen2017hdr, hdrunet_chen2021hdrunet}.
Applying element-wise addition to combine features from different layers can reduce computation. However, it can cause conflicts as one feature represents information with a large receptive field, while the other may represent high-dimensional features. 
A direct solution is to introduce attention mechanisms at a shallow level to extract salient features. However, existing methods like SE\cite{hu2018senet} and ECA\cite{wang2020eca} solely obtain attention weights from the encoder, which may not align with the decoder's requirements.

To address this challenge, we propose Skip-Channel Attention (SCA). 
What sets SCA apart is its ability to compute salient features by incorporating guidance from the decoder. 
As shown in Fig.\ref{fig:fun-net}[b], we compute global spatial information at both shallow and deep layers, merging them to create a more comprehensive result.
A two-layer MLP is then utilized to capture channel-wise relationships and generate an attention vector. 
Subsequently, each channel in the shallow features is multiplied by the corresponding elements in the attention vector and then added to the deep features.
Compared to the original naive skip-connection mechanism, our SCA requires only a minimal increase in computational cost while significantly enhancing the ability to mitigate conflicts across various layers.

\subsection{Siamese Self-Attention Block (SSAB)}\label{SSAB}
Fig.\ref{fig:fun-net}[c] illustrates the Siamese Self-Attention Block (SSAB).
Note that the attention mechanism in SSAB does not rely on query-key pairwise interactions. 

Starting with a layer-normalized tensor $X$, we apply $1\times1$ convolutions and $3\times3$ depth-wise convolutions to aggregate channel and local context.
The resulting feature maps are then split into two equal segments, denoted as $Q\in R^{N*h*(C/h)}$ and $P\in R^{N*h*(C/h)}$, representing a sequence of $N$ feature vectors with feature size $C/h$ and head size $h$.
These two segments are referred to as siameses, and serve the purpose of computing self-attention weight and subsequent complementary fusion.
To calculate the self-attention weight, we propose two different gating mechanisms in SSAB. 
The gate in the first part focuses on the complementary effects of siamese features, while the gate in the second part are designed to add nonlinear mapping capabilities.

In the first part of our approach ($SSA_{1}$), we project $Q$ through a saliency-gate matrix $W\in R^{h*(C/h)}$ and normalize it across all positions using softmax to generate attention map $A\in R^{N*h}$:

\begin{equation}
	A = softmax(\frac{QW}{\sqrt{D}})
\end{equation}

Where $D$ is a scaling factor. And then, we perform a weighted accumulation of $Q$ to produces multi-head global context:

\begin{equation}
	G = \sum_{i=1}^{N} (A_{i}  Q_{i})
\end{equation}

The global context undergoes element-wise multiplication with $P$ to build local-global interactions:

\begin{equation}
	Y_{1} = ((G \odot P) L_{1}) + Q ) L_{2}
\end{equation}

Where $\odot$ denotes the element-wise product, $L_{1}$ and $L_{2}$ are linear projection matrices. We employ local residual learning to obtain the final result of the first part:

\begin{equation}
	SSA_{1}(X) = X + Y_{1}
\end{equation}

In the second part of our approach ($SSA_{2}$), we aim to incorporate a non-linear mapping to enhance the non-linearity of SSAB. 
We follow a similar procedure as in the first part to initially derive siamese components.
Subsequently, we employ the gelu function as a gate mechanism on siamese $Q$ to generate attention weights, which are then integrated with another siamese $P$:

\begin{equation}
	Y_{2} = Gelu(Q) \odot  P
\end{equation}

Local residual learning is also employed in $SSA2$:

\begin{equation}
	SSA_{2}(X) = X + Y_{2}
\end{equation}

The entire process can be modeled as:

\begin{equation}
	SSAB = SSA_{1}(X) + SSA_{2}(SSA_{1}(X))
\end{equation}

VIT compute elements' position-to-position pairwise interactions, resulting in quadratic computational complexity. SSAB is fundamentally different from VIT, employing local-to-global interactions to replace these pairwise interactions to achieve linear computational complexity.
The concatenation of $SSA1$ and $SSA2$ is critical, as both components playing indispensable roles in the functionality of SSAB.

\subsection{Two-Stage Cascade Framework}
The preprocessing pipeline for the SID dataset involves two key steps: subtracting black levels and applying amplification at desired ratios (x100, x250, and x300)\cite{sid_chen2018_sid}.
Despite the widespread adoption of this same preprocessing method, introduced by SID in 2018, the underlying issues have not been comprehensively investigated. 
Based on our observations, although the amplification factor reduces the amplitude gap between short and long exposure data, it also increases noise in the short-exposure input data, posing challenges in the subsequent image restoration process.
We can model the noise characteristics of the raw image as a combination of Poisson noise and Gaussian noise, as shown below:
\begin{equation}\label{eq_noiseModel}
	x \sim k \mathcal{P}\left(\frac{x^{*}}k\right)+\mathcal{N}\left(0, \sigma ^{2}\right)
\end{equation}

Where $x^{*}$ is the pixel value from expected photons, $k$ is the product of system gain and quantum efficiency factor. $\mathcal{N}$ is the sum of Gaussian noise before and after applying system gain. 
When we amplify the short-exposure data using a certain ratio, it is inevitable to amplify the noise contained in the image at the same time:

\begin{equation}
	x_{amp} \sim k \mathcal{P}\left(\frac{x^{*}}k\right) ratio + \mathcal{N}\left(0, \sigma ^{2}\right) ratio
\end{equation}

The SID dataset ensures that the camera's aperture, ISO, focus, and focal length remain constant in each scene. The only variable is the exposure time. Consequently, the long-exposure reference images can be expressed as follows:

\begin{equation}
	x_{ref} \sim k \mathcal{P}\left(\frac{x^{*} ratio}k\right)+\mathcal{N}\left(0, \sigma ^{2}\right)
\end{equation}

To analyze this distribution, a usual simplification is to treat the Poisson
distribution $P(\lambda )$ as a Gaussian distribution of $N(\lambda ,\lambda)$:

\begin{equation}\label{eq_compare}
	\begin{aligned}
		x_{amp} 
		&\sim k \mathcal{N}\left(\frac{x^{*}ratio}{k}, \frac{x^{*}ratio ^{2}}{k}\right)+ \mathcal{N}\left(0, \sigma ^{2} ratio^{2}\right)
		\\
		x_{ref} 
		&\sim k \mathcal{N}\left(\frac{x^{*}}{k} ratio , \frac{x^{*}}{k} ratio \right) + \mathcal{N}\left(0, \sigma ^{2}\right) 
	\end{aligned}
\end{equation}

As indicated in Eqn.(\ref{eq_compare}),  it is evident that the amplified short-exposure data and long-exposure data exhibit distinct noise levels 
The noise is magnified by a factor of approximately $ratio$, which will seriously hinder the restoration of the image. To eliminate the amplified noise introduced by the preprocessing process, we adopt a two-stage cascade framework. This framework comprises raw image denoising and the transformation of raw data to sRGB (raw2rgb). For both stages, we utilize the same architecture.

To train the denoising ability of the first-stage network, we employ long-exposure data as reference images and generate corresponding simulated noise to obtain degraded images.  
To achieve this, we require precise estimation of $k$ and $\sigma ^{2}$ under a particular sensor's ISO setting.
Fortunately, the mean and variance over $x$ can be transformed into the following linear regression problem\cite{wang2020practical}:

\begin{equation}\label{eq_mean_var}
	\begin{aligned}
		&E(x)=x^{*}\\ 
		&Var(x)=kx^{*}+\sigma ^{2}
	\end{aligned}
\end{equation}

Thus, we captured a sequence of raw images of a static gray-scale chart to sample the real-world noise. 
Subsequent to training the first-stage denoising model, we integrate it with the second-stage raw2rgb model and conduct fine-tuning for the final result.

We choose to adopt a two-stage approach for the following reasons: 
(1) The amplified short-exposure data matches the ground truth in numerical amplitudes but is marred in noisy distribution. Training with contaminated data would significantly increase network learning difficulty.
(2) Denoising is more feasible to execute on raw data, while restoring brightness is more achievable with clean raw data. Decoupling serves as a solution to address domain conflict issues.
(3) The two stages are relatively decoupled, ensuring that local network adjustments do not impact downstream tasks and facilitates easier convergence of the overall framework. 


\section{Experiments and Results}
\label{sec:Experiments}

\subsection{Experimental Setup}
{\bf Dataset.} We adopt the extreme low-light SID dataset\cite{sid_chen2018_sid} as the benchmark, consisting of 5094 raw short-exposure images captured from Sony and Fuji sensors. 
Sony dataset use Bayer sensor in the image resolution of 4256 $\times$ 2848 and Fuji dataset use X-Trans sensor with 6032 $\times$ 4032 image resolution.
Noting misalignment in long-short pairs of three scenes in the test set of Sony subsets, we exclude these images in testing following prior method\cite{did_maharjan2019did}.

{\bf Implementation Details.}  
The networks are trained from scratch using L1 and MS-SSIM loss functions and the Adam optimizer for 4000 epochs. 
During training, a random 512$\times$512 patch was cropped from the original raw image with horizontal and vertical flipping, and the batch size was set to 1.

\subsection{Results}

\begin{table*}
	\begin{center}
		\setlength{\tabcolsep}{2mm}{
			\begin{tabular}{cccccccc}
				\hline
				\multirow{2}{*}{Method} & 
				\multirow{1}{*}{PSNR ($dB\uparrow$)}& 
				\multirow{1}{*}{SSIM ($\uparrow$)}&
				\multirow{1}{*}{PSNR ($dB\uparrow$)}& 
				\multirow{1}{*}{SSIM ($\uparrow$)}& \multirow{2}{*}{\#GMACs ($\downarrow$)} & \multirow{2}{*}{\#Params($M\downarrow$)}\\
				\cline{2-5}
				& \multicolumn{2}{c}{Sony} & \multicolumn{2}{c}{Fuji} &  &  \\
				
				\hline
				SID\cite{sid_chen2018_sid} (18, CVPR) & 28.88 & 0.787 & 26.61 & 0.680 & 562.06  & 7.7 \\
				
				DID\cite{did_maharjan2019did} (19, ICME) & 28.41 & 0.780 & - & - & $>$2000  &  2.5 \\
				
				SGN\cite{sgn2019gu} (19, ICCV) & 28.91 & 0.789 & {\bf{\color{red}26.90}} & {\bf{\color{red}0.683}} & $>$2000  &  3.5 \\
				
				LDC@\cite{lcd_xu2020lcd} (20, CVPR) & 29.56 & {\bf{\color{green}0.799}} & 26.70 & 0.681 & $>$2000 & 8.6 \\
				
				LLPackNet*\cite{llpack_lamba2020LLPackNet} (20, BMVC) & 27.83 & 0.750 & 24.13 & 0.59 & 83.46  &  1.16 \\
				
				RED*\cite{red_lambaCvpr2021Red} (21, CVPR) & 28.66 & 0.790 & 26.60 & 0.682 & {\bf{\color{yellow}59.8}} & {\bf{\color{yellow}0.78}}  \\
				
				DBLE* \cite{abandoning_dong2022abandoning} (22, CVPR) & {\bf{\color{red}29.65}} & {\bf{\color{red}0.797}} & - & - & 1162.51 & 15.01  \\
				
				Ours (one-stage) & {\bf{\color{yellow}30.16}} & {0.795} & {\bf{\color{yellow}27.75}} & {\bf{\color{yellow}0.717}} & {\bf{\color{green}34.97}} & {\bf{\color{green}0.46}} \\
				
				Ours (two-stage) & {\bf{\color{green}30.55}} & {\bf{\color{yellow}0.798}} & {\bf{\color{green}27.86}}$^{\diamond}$ & {\bf{\color{green}0.718}}$^{\diamond}$ & {\bf{\color{red}63.83}} & {\bf{\color{red}0.92}}  \\
				\hline
		\end{tabular}}
	\end{center}
	\caption{Quantitative comparisons to the SOTA methods. The best, second, and third results are marked in {\bf{\color{green} green}}, {\bf{\color{yellow} yellow}}, and {\bf{\color{red} red}} perspectively. \#Params and \#GMACs represent the number of model parameters and Giga Multiply-Accumulate Operations. \#GMACs is measured in the Sony dataset. Methods with @ indicate that the model is trained and tested on down-sampled images. Methods with * indicate that the model is trained and tested on low digital gain images(x100). $^{\diamond}$ indicate that the model is trained with random noise due to camera limitations.} \label{table:all}
\end{table*}

We compare our results with several recent low-light raw image enhancement methods, including SID\cite{sid_chen2018_sid}, DID\cite{did_maharjan2019did}, LLPackNet\cite{llpack_lamba2020LLPackNet}, LCD\cite{lcd_xu2020lcd}, RED\cite{red_lambaCvpr2021Red}, SGN\cite{sgn2019gu} and DBLE\cite{abandoning_dong2022abandoning}. 

In Table \ref{table:all}, we evaluate our approach using two widely-used quantitative assessment methods: peak signal-to-noise ratio (PSNR) and structural similarity (SSIM).
However, several efficiency metrics are also considered in evaluation for real-world application, such as the number of multiply-accumulate (MAC) operations and parameters.
We can see our proposed method outperforms other state-of-the-art methods by a significant margin across various metrics. 
Note that some of previous methods\cite{lcd_xu2020lcd, llpack_lamba2020LLPackNet, red_lambaCvpr2021Red, abandoning_dong2022abandoning} were trained and tested under different schemes, which might result in better performance\footnote{ LDC\cite{lcd_xu2020lcd} is trained and tested on down-sampled images, while LLPackNet\cite{llpack_lamba2020LLPackNet}, RED\cite{red_lambaCvpr2021Red}, and DBLE\cite{abandoning_dong2022abandoning} only train and test images with low digital gains (x100).}
Specifically, when compared to the state-of-the-art method on Sony and Fuji datasets, our two-stage approach achieved PSNR improvements of 0.9 dB and 0.96 dB, respectively.

\begin{figure}[htbp]
	\centering	
	\subfigure[Input Raw]{
		\begin{minipage}[t]{0.24\columnwidth}
			\centering
			\includegraphics[width=0.98\columnwidth]{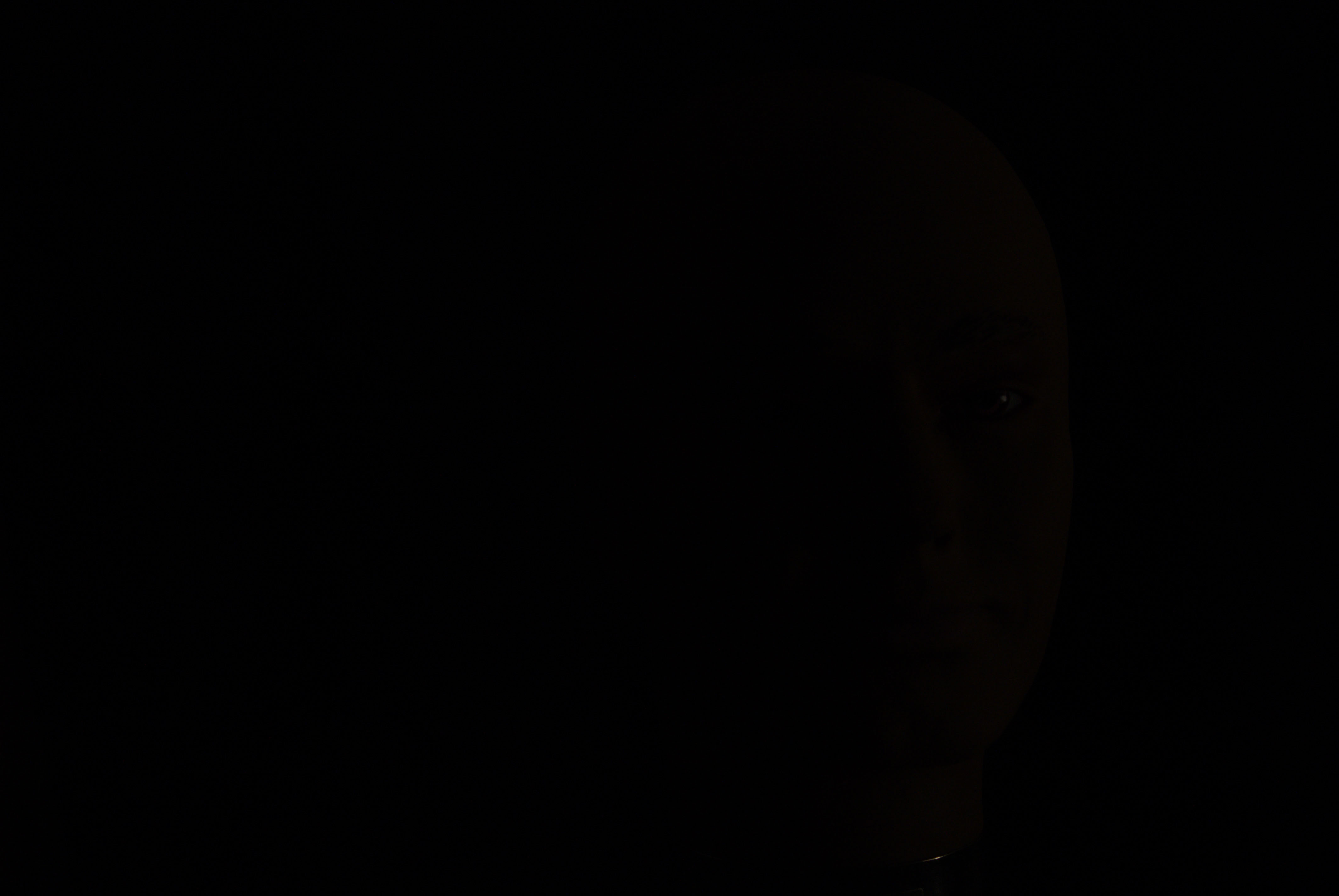}
		\end{minipage}%
	}%
	\subfigure[SID\cite{sid_chen2018_sid}]{
		\begin{minipage}[t]{0.24\columnwidth}
			\centering
			\includegraphics[width=0.98\columnwidth]{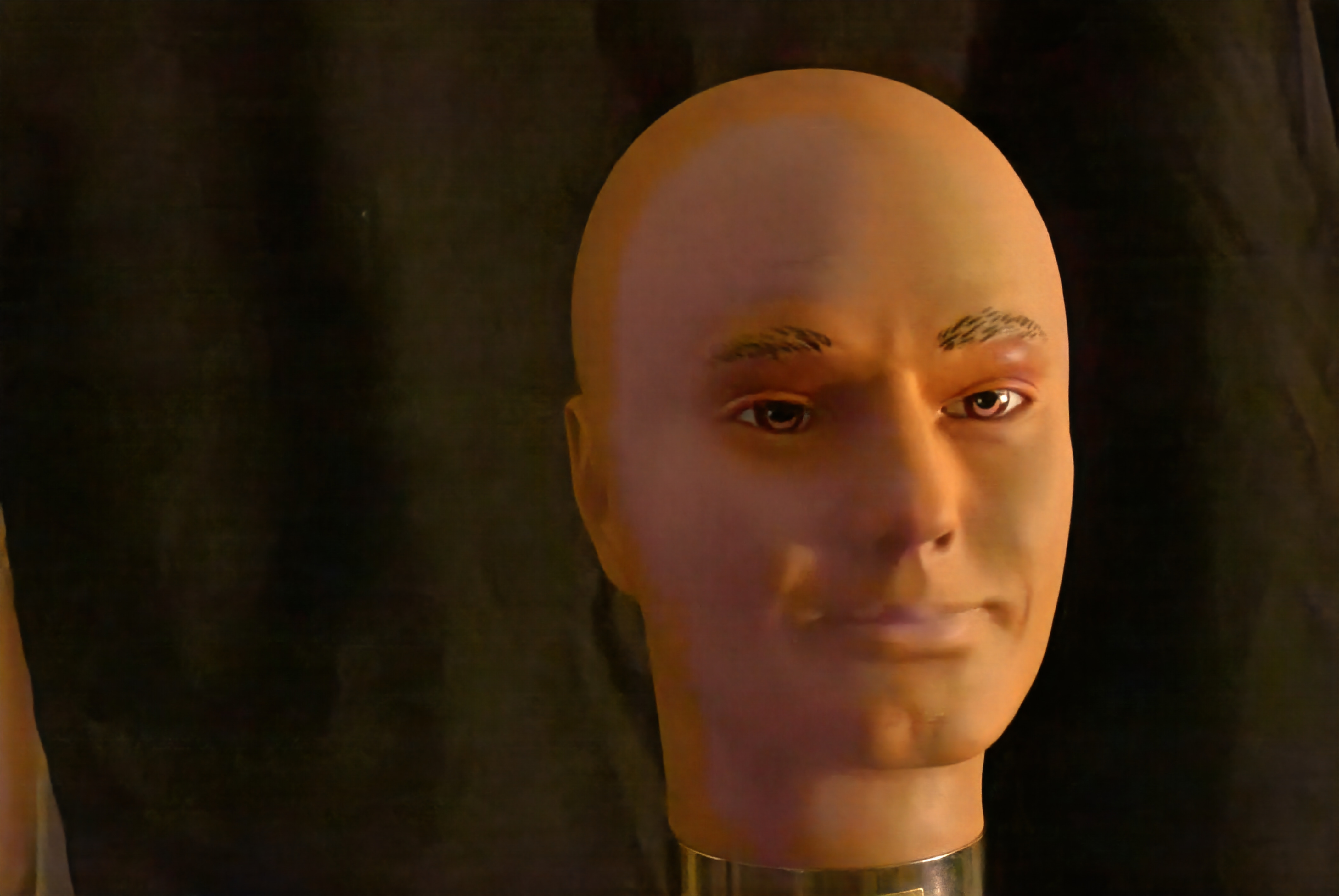}
		\end{minipage}%
	}%
	\subfigure[DID\cite{did_maharjan2019did}]{
		\begin{minipage}[t]{0.24\columnwidth}
			\centering
			\includegraphics[width=0.98\columnwidth]{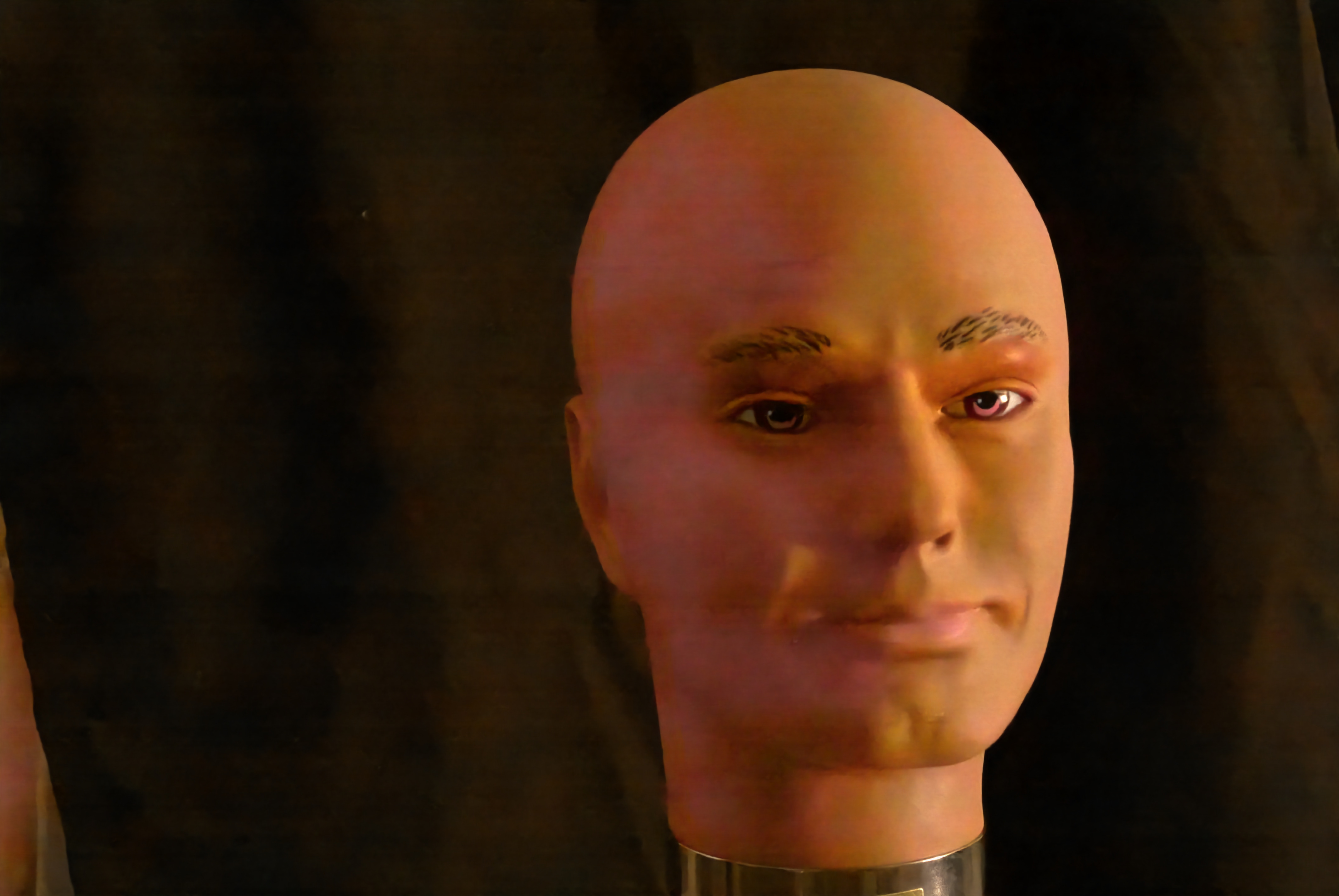}
		\end{minipage}%
	}%
	\subfigure[RED\cite{red_lambaCvpr2021Red}]{
		\begin{minipage}[t]{0.24\columnwidth}
			\centering
			\includegraphics[width=0.98\columnwidth]{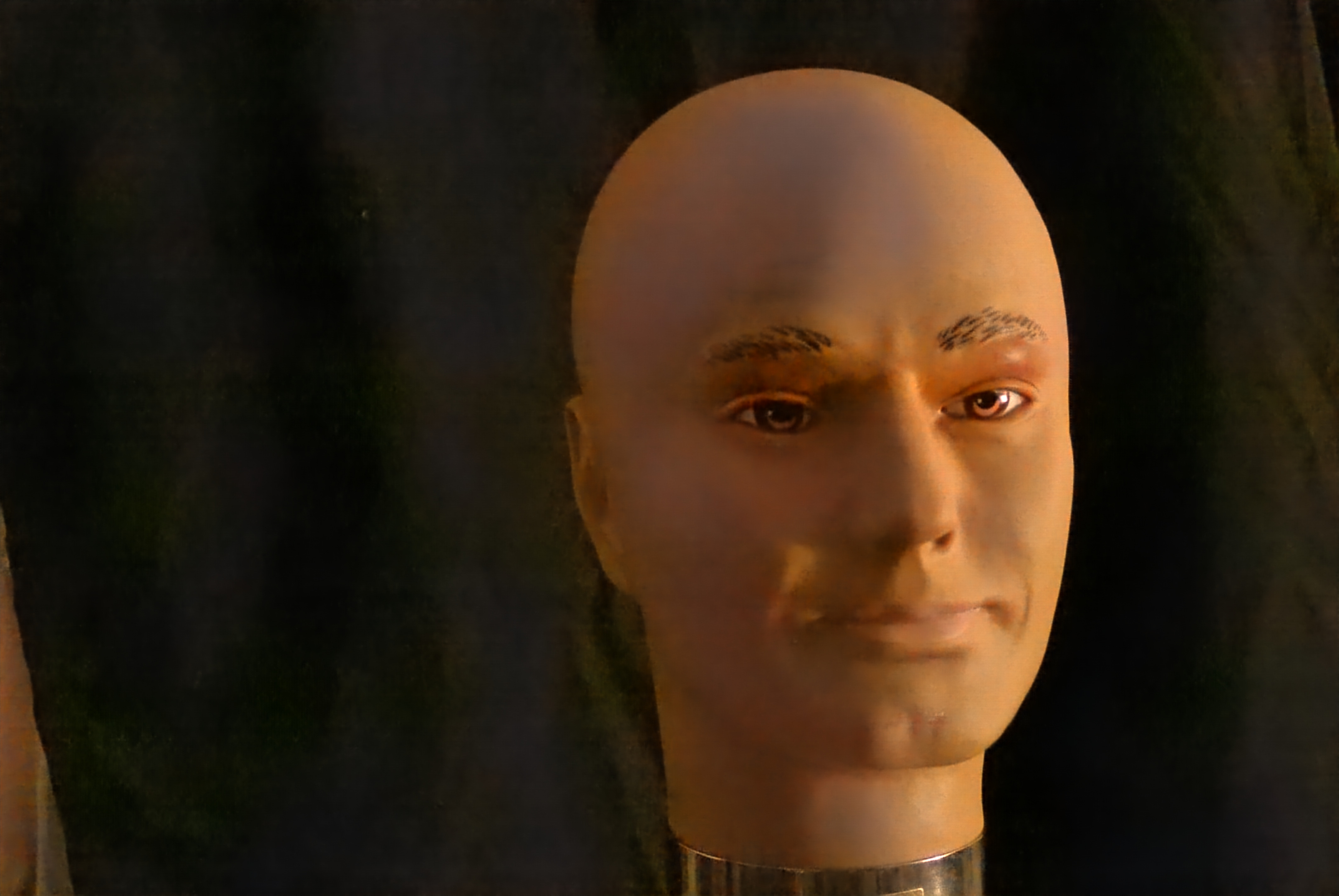}
		\end{minipage}%
	}%
	\quad
	\subfigure[DBLE\cite{abandoning_dong2022abandoning}]{
		\begin{minipage}[t]{0.24\columnwidth}
			\centering
			\includegraphics[width=0.98\columnwidth]{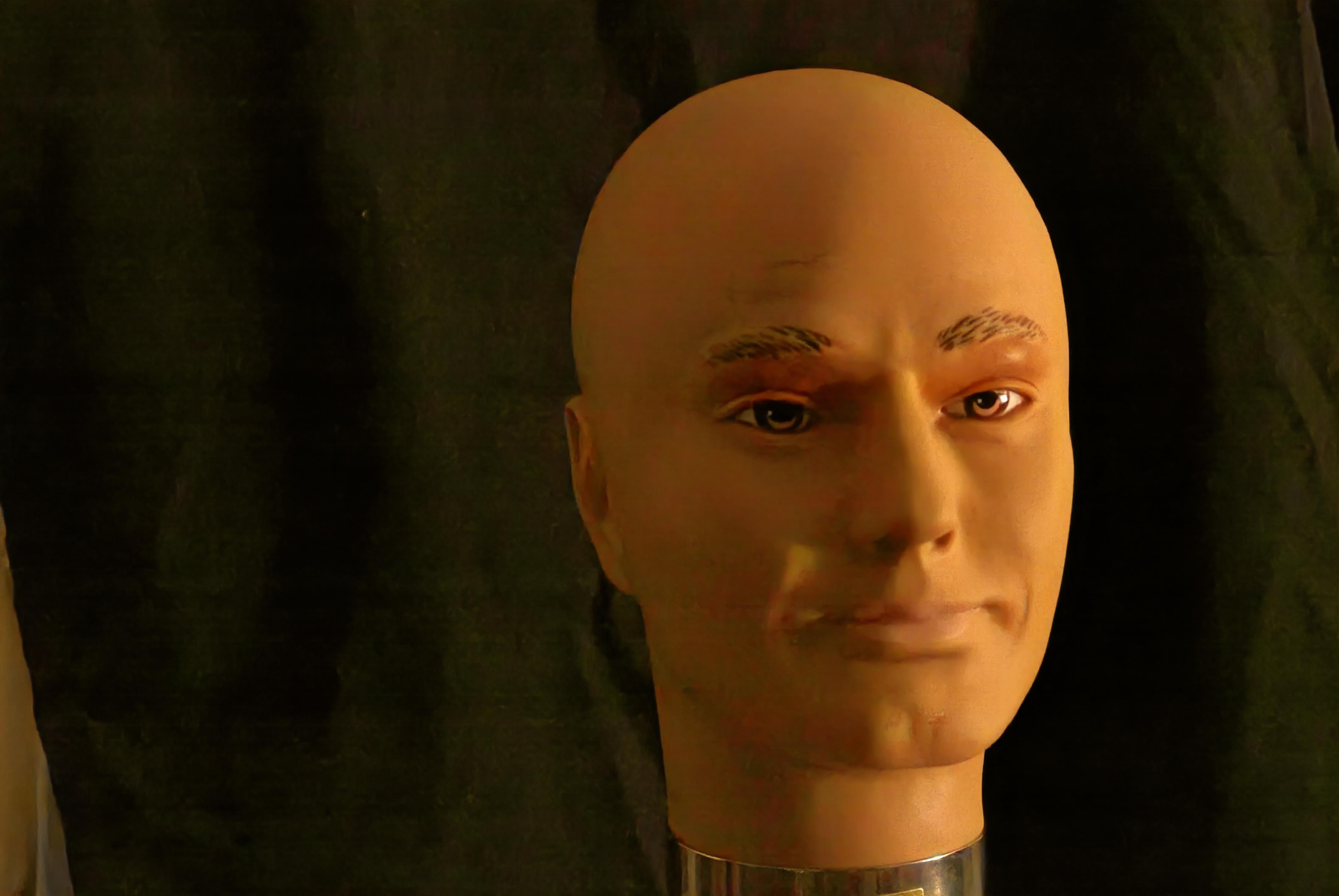}
		\end{minipage}%
	}%
	\subfigure[Ours(1-stage)]{
		\begin{minipage}[t]{0.24\columnwidth}
			\centering
			\includegraphics[width=0.98\columnwidth]{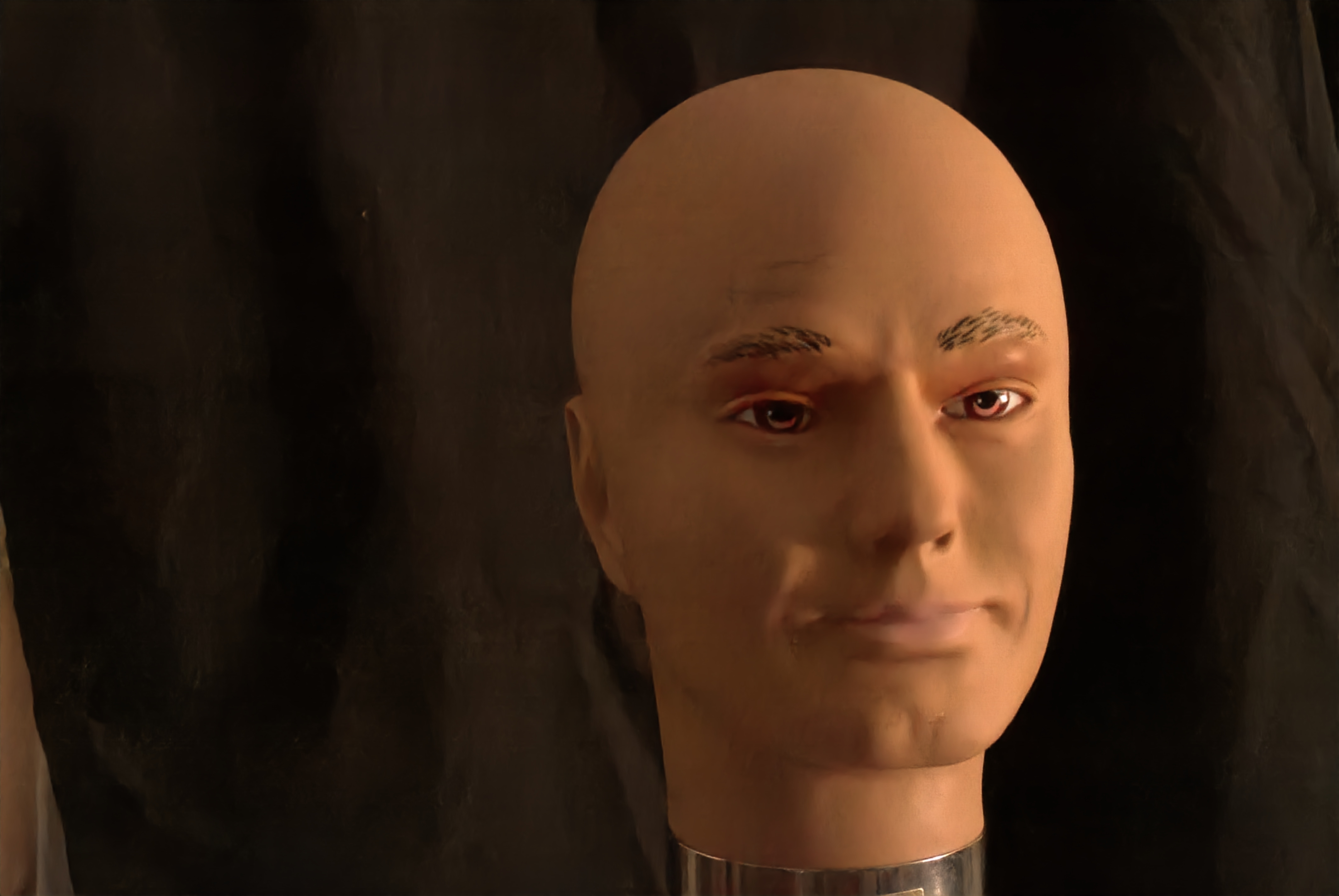}
		\end{minipage}%
	}%
	\subfigure[Ours(2-stage)]{
		\begin{minipage}[t]{0.24\columnwidth}
			\centering
			\includegraphics[width=0.98\columnwidth]{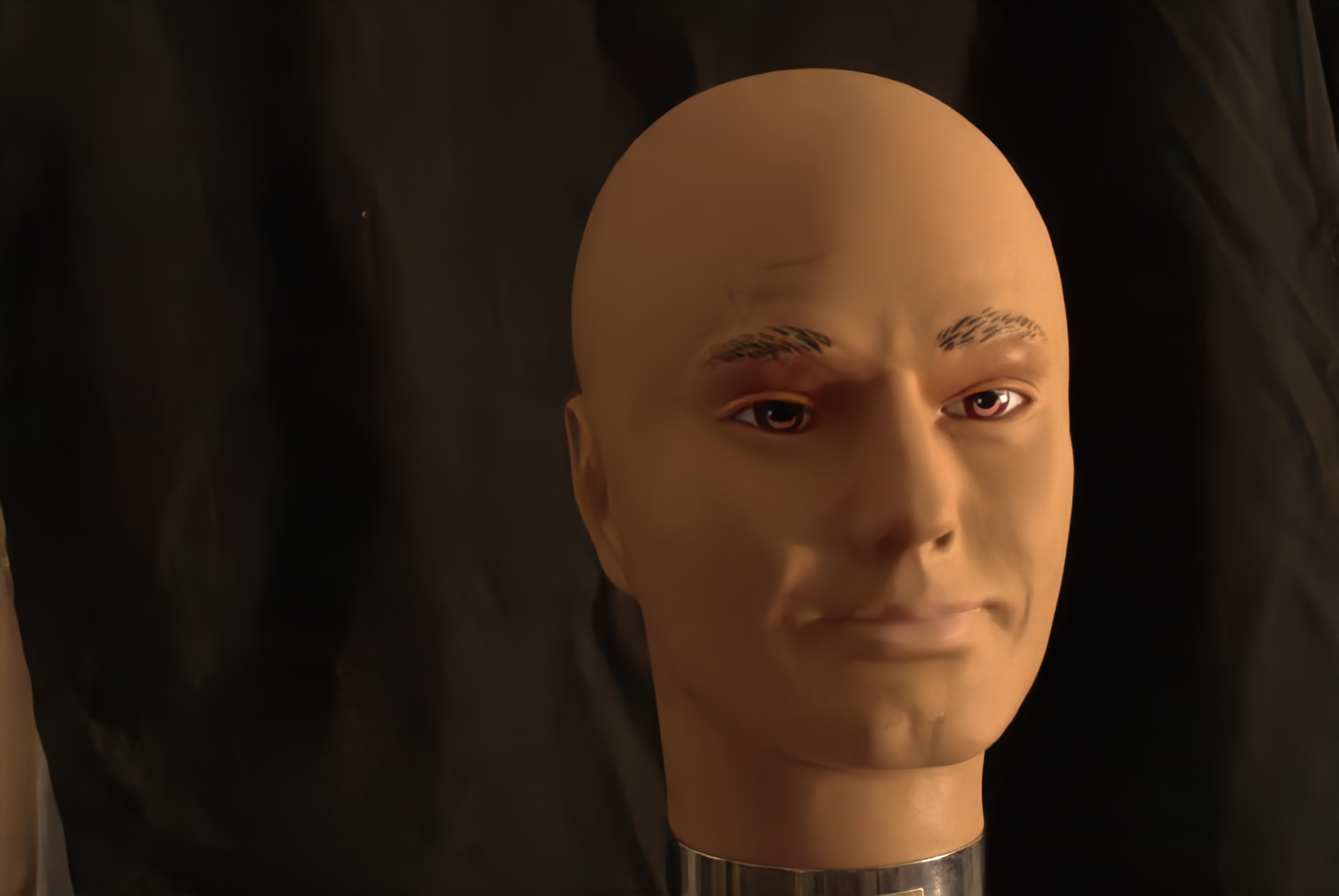}
		\end{minipage}%
	}%
	\subfigure[GT]{
		\begin{minipage}[t]{0.24\columnwidth}
			\centering
			\includegraphics[width=0.98\columnwidth]{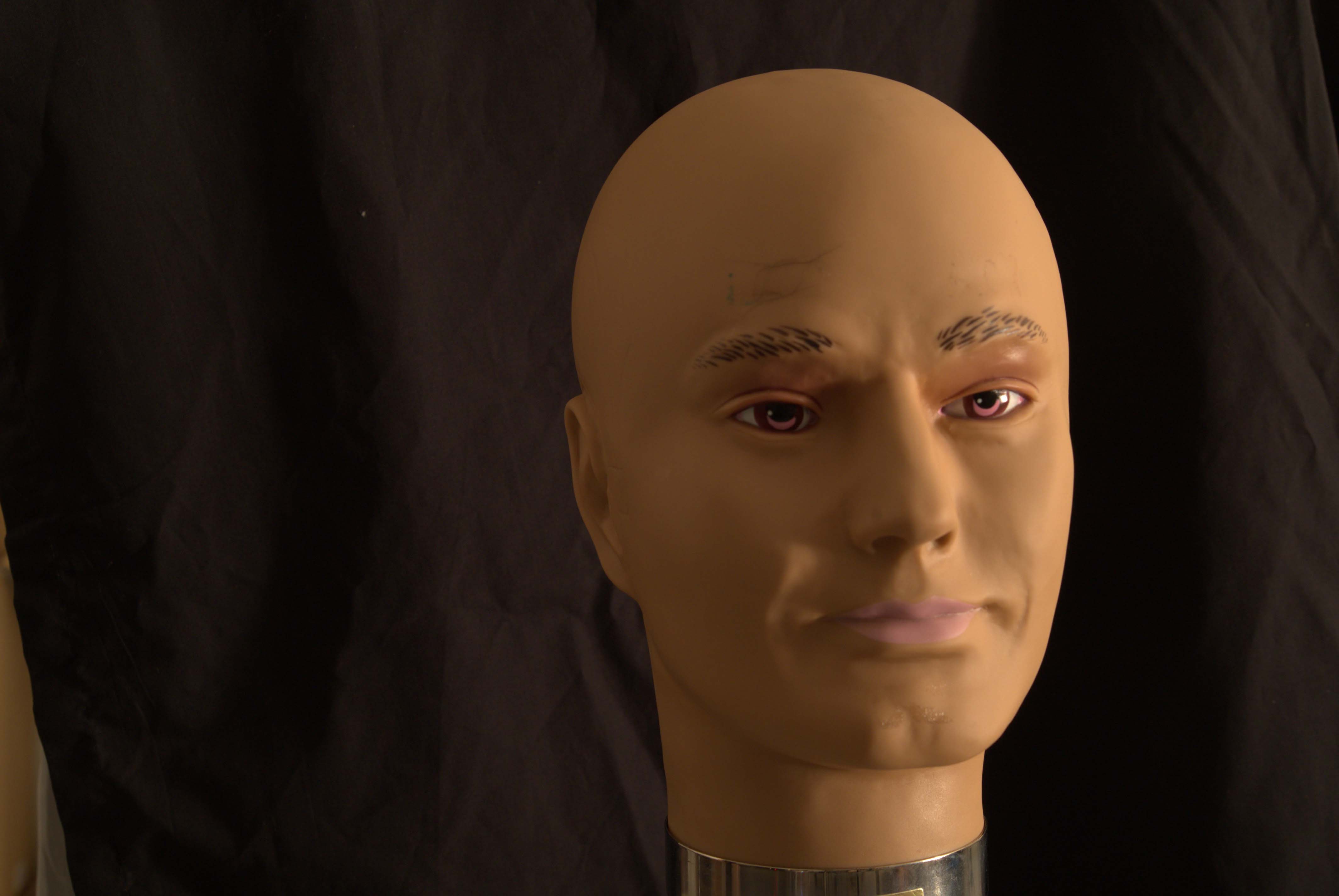}
		\end{minipage}%
	}%
	\centering
	
	\caption{Qualitative Comparison of ours and previous state-of-the-art methods on the SID Sony dataset.}\label{fig:compare_with_sota}
\end{figure}

The visual comparison presented in Figure \ref{fig:compare_with_sota} highlights the superiority of our models in terms of color consistency, noise reduction, and detail preservation, outperforming the compared methods, which display differing levels of color distortion and noise.
Notably, the two-stage model exhibits lower noise levels than the one-stage model, confirming the effectiveness of our two-stage framework.

\subsection{ Ablation Studies}\label{ablation}
We base the individual components of our proposed method to make ablation studies. The base network is a sole Unet-based network. 

\begin{table}
	\begin{center}
		\begin{tabular}{ccc}
			\hline
			Method & PSNR & SSIM\\
			\hline
			Base + SCA & 25.68 & 0.744 \\	
			Base + SCA + SSA1 & 29.19 & 0.783 \\
			Base + SCA + SSA2 & 29.44 & 0.789 \\
			Base + SCA + SSA1 + SSA1 & 29.30 & 0.783 \\
			Base + SCA + SSA2 + SSA2 & 29.54  & 0.789 \\
			Base + SCA + SSA1 + SSA2 & \bf 30.16 & \bf 0.795 \\
			\hline
		\end{tabular}
	\end{center}
	\caption{Ablation studies of our proposed SSAB on  Sony dataset.} \label{table:ssa}
\end{table}

{\bf Siamese Self-Attention Block.} 
The results presented in Table \ref{table:ssa} indicate that incorporating either $SSA_{1}$ or $SSA_{2}$ component leads to improved performance.
Furthermore, as detailed in Section \ref{SSAB}, $SSA_{1}$ and $SSA_{2}$ serve distinct roles, and their combination can yield optimal performance.

\begin{table}
	\begin{center}
		\begin{tabular}{ccc}
			\hline
			Method & PSNR & SSIM\\
			\hline
			Base + SSAB & 29.96 & 0.792 \\
			Base + SSAB + SE Attention\cite{hu2018senet} & 30.02 & 0.793 \\
			Base + SSAB + ECA Attention\cite{wang2020eca} & 30.02  & 0.793 \\
			Base + SSAB + SCA Attention & \bf 30.16 & \bf 0.795 \\
			\hline
		\end{tabular}
	\end{center}
	\caption{Comparisons of SCA and other methods on Sony dataset.} \label{table:sca}
\end{table}

{\bf Skip-Channel Attention.}
We assessed the performance of our proposed SCA by comparing it with the widely used channel attention mechanisms SENet\cite{hu2018senet} and ECA\cite{wang2020eca}.
SE, ECA, and SCA were all applied at the same position in our evaluation. SE and ECA operated on the encoder following their original implementations to extract salient features. 
In contrast, SCA diverges in that it computes salient features by incorporating guidance from the decoder.
As illustrated in Table \ref{table:sca}, our SCA surpasses SE and ECA attention, demonstrating the effectiveness of our approach.

{\bf Two-stage Pipeline.} 
As shown in the lower rows of Table \ref{table:all}, our two-stage model outperforms the one-stage model by 0.39dB PSNR in the Sony dataset. 
We also evaluated the performance of the two-stage model trained with random noise, as detailed in Table \ref{table:two-stage}, which revealed a PSNR reduction of 0.3 dB.
Training the two-stage model directly did not yield significantly improved results compared to the one-stage model.
As the two-stage framework is less effective without real-noise guided training.

\begin{table}
	\begin{center}
		\begin{tabular}{ccc}
			\hline
			Method & PSNR & SSIM\\
			\hline
			Ours (one-stage) & 30.16 & 0.795 \\
			Ours (two-stage) + w/o noise & 30.21 & 0.795 \\
			Ours (two-stage) + random noise & 30.25 & 0.796 \\
			Ours (two-stage) + real noise & \bf 30.55 & \bf 0.798 \\
			\hline
		\end{tabular}
	\end{center}
	\caption{Ablation studies of our two-stage pipeline on Sony dataset.} \label{table:two-stage}
\end{table}


\section{Conclusion}
\label{sec:Conclusion}

In this paper, we propose a lightweight deep learning-based solution for extreme low-light raw image restoration. 
To address the challenge of high computational demands associated with self-attention, we propose Siamese Self-Attention Block (SSAB), whose computational complexity grows linearly with the input size. 
Leveraging SSAB and our Skip-Channel Attention (SCA), a lightweight network is proposed, which is superior to the existing state-of-the-art (SOTA) methods with the cost of minimal computational resources. 
Furthermore, we designed a Two-Stage Framework to solve the long-standing domain conflict problem and achieved superior results than single-stage models. 
Both comparative experiments and ablation studies validate the effectiveness of our proposed method.

\bibliographystyle{IEEEbib}
\bibliography{strings,refs}

\end{document}